\documentclass[journal]{IEEEtran}

\usepackage{amsmath}
\usepackage{mathrsfs}
\usepackage{amsfonts}
\usepackage{threeparttable}
\usepackage{changes}
  \definechangesauthor[name={Per cusse}, color=orange]{per}
\usepackage{amsmath,amssymb}
\usepackage{acro}
\usepackage{multirow}
\usepackage{cite}
\usepackage{setspace}
\usepackage{verbatim}
\usepackage{soul}
\soulregister\cite7 
\soulregister\citep7 
\soulregister\citet7 
\soulregister\ref7 
\soulregister\pageref7 
\usepackage{bm}
\usepackage[ruled]{algorithm2e}
\usepackage{algorithmic}
\usepackage{float}

\usepackage{amsmath}

\ifCLASSOPTIONcompsoc
  \usepackage[caption=false,font=normalsize,labelfont=sf,textfont=sf]{subfig}
\else
  \usepackage[caption=false,font=footnotesize]{subfig}
\fi
\usepackage{url}
\usepackage{changepage}
\usepackage{color} 
\usepackage{xcolor}
\usepackage[normalem]{ulem} 
\newcommand\phaseA{\bgroup\markoverwith
  {\textcolor{blue!20}{\rule[-.5ex]{2pt}{2.5ex}}}\ULon}
\newcommand\phaseB{\bgroup\markoverwith
  {\textcolor{green!20}{\rule[-.5ex]{2pt}{2.5ex}}}\ULon}
\newcommand\phaseC{\bgroup\markoverwith
  {\textcolor{red!20}{\rule[-.5ex]{2pt}{2.5ex}}}\ULon}
\usepackage{framed}
\usepackage{adjustbox}

\usepackage{nomencl}

\usepackage[hang,flushmargin]{footmisc}
\usepackage{lipsum}
\makeatletter
\newcommand{\algorithmfootnote}[2][\footnotesize]{%
  \let\old@algocf@finish\@algocf@finish
  \def\@algocf@finish{\old@algocf@finish
    \leavevmode\rlap{\begin{minipage}{\linewidth}
    #1#2
    \end{minipage}}%
  }%
}
\makeatother

\usepackage{amsmath}

\DeclareSymbolFont{boldoperators}{OT1}{cmr}{bx}{n}
\SetSymbolFont{boldoperators}{bold}{OT1}{cmr}{bx}{n}
\edef\bar{\unexpanded{\protect\mathaccentV{bar}}\number\symboldoperators16}

\makenomenclature

\begin{document}
%
\title{Enhancing Short-Term Wind Speed Forecasting using Graph Attention and Frequency-Enhanced Mechanisms}

%
%
%

\author{Hao~Liu, 
        Huimin~Ma,~\IEEEmembership{Member,~IEEE,}
        Tianyu~Hu,~\IEEEmembership{Member,~IEEE,}
\thanks{H. Liu, T. Hu and H. Ma are with the School of Computer and Communication Engineering, University of Science and Technology Beijing, Beijing, China.}
}

%
%

\maketitle

\begin{abstract}
The safe and stable operation of power systems is greatly challenged by the high variability and randomness of wind power in large-scale wind-power-integrated grids. Wind power forecasting is an effective solution to tackle this issue, with wind speed forecasting being an essential aspect. In this paper, a Graph-attentive Frequency-enhanced Spatial-Temporal Wind Speed Forecasting model based on graph attention and frequency-enhanced mechanisms, i.e., GFST-WSF, is proposed  to improve the accuracy of short-term wind speed forecasting. The GFST-WSF comprises a Transformer architecture for temporal feature extraction and a Graph Attention Network (GAT) for spatial feature extraction. The GAT is specifically designed to capture the complex spatial dependencies among wind speed stations to effectively aggregate information from neighboring nodes in the graph, thus enhancing the spatial representation of the data. To model the time lag in wind speed correlation between adjacent wind farms caused by geographical factors, a dynamic complex adjacency matrix is formulated and utilized by the GAT. Benefiting from the effective spatio-temporal feature extraction and the deep architecture of the Transformer, the GFST-WSF outperforms other baselines in wind speed forecasting for the 6-24 hours ahead forecast horizon in case studies.
\end{abstract}


\IEEEpeerreviewmaketitle

\section{Introduction}

\indent In recent years, the rapid development of modern industry and the exponential growth of population have caused environmental pollution and global energy crisis to become increasingly serious. As a result, renewable energy has gained widespread attention for its potential to solve environmental and energy problems. Among clean renewable energy sources, wind power has seen overwhelming growth over the past decade, with global wind power capacity reaching 837 GW in 2021 \cite{council2022global}. However, the variability of wind speed can lead to intermittent power generation, which poses challenges to the safety and stability of smart grid energy management systems. Studies have shown that wind power forecasting is one of the most cost-effective and efficient methods to minimize this risk \cite{wind_power_review2011,wang2017deep,wang2021review}. Therefore, accurate wind power forecasting is urgently needed, as well as wind speed forecasting, which actually serves as the foundation for wind power forecasting \cite{2014review, 2020probabilistic}.

\indent Numerous wind speed forecasting methods have been proposed, including physical, statistical, machine learning, and deep learning methods. Physical methods mostly rely on Numeric Weather Prediction (NWP), which uses weather data such as temperatures and pressure to solve complex mathematical and physical models and forecast wind speed \cite{NWP}. Statistical methods, on the other hand, construct a linear or nonlinear function from historical data to forecast future wind speed values \cite{wang2016analysis}, e.g., Auto-Regressive Moving Average (ARMA) \cite{ARMA}, Auto-Regressive Integrated Moving Average (ARIMA) \cite{ARIMA}, and Eupport Vector Regression (SVR) \cite{SVR}, etc. 

\indent Artificial intelligence technologies have led to the emergence and development of numerous machine learning and deep learning models for wind speed forecasting. AI-based short-term wind power forecasting methods include Extreme Learning Machines (ELM) \cite{ELM}, Light Gradient Boosting Machine (LightGBM) \cite{2017lightgbm}, Artificial Neural Network (ANN) \cite{ANN}, Convolutional Neural Network (CNN)  \cite{CNN}, Recurrent Neural Network (RNN) \cite{RNN}, Long Short-term Memory Network (LSTM) \cite{hu2018LSTM}, among others. However, these methods have mainly relied on historical data from individual wind farms, hardly considering cross-farm spatial correlation \cite{hu2019very}. As multiple wind farms close to each other are typically within the same wind belt, wind speeds are strongly correlated spatially. As a result, several studies considering both temporal and spatial correlation between adjacent wind fields have emerged in recent years, and have further improved forecasting performance \cite{2014graph-Lstm, 2018CNN-MLP,2019CNN-LSTM,2020GCN-LSTM}.

\indent In \cite{2014graph-Lstm}, a spatial vector consisting of wind speed values from multiple nodes was used as the spatial feature input to the forecasting model, while the temporal features were extracted by LSTM. \cite{2018CNN-MLP} proposed a model for wind speed forecasting with spatio-temporal correlation, integrating CNN and a multi-layer perceptron (MLP). Here, the spatial features were extracted by CNN, and the MLP captured the temporal dependencies among these extracted spatial features. In \cite{2019CNN-LSTM}, a deep architecture named the Predictive Spatio-Temporal Network (PSTN) was used for wind speed forecasting, integrating CNN and LSTM. Similar to the CNN-MLP model, the temporal feature extraction module of this model was changed from MLP to LSTM. Finally, in \cite{2020GCN-LSTM}, a model was proposed to forecast the wind farm cluster by combining Graph Convolution Networks (GCN) with LSTM. The performance improvements observed in these methods, whether using CNN or Graph Neural Network (GNN), demonstrate the effectiveness of utilizing spatial correlation for wind speed forecasting. Therefore, we utilize the GAT\cite{2017GAT} to extract spatial correlation features between neighboring wind farms, aiming to further enhance the forecasting performance of our model.


\indent Recently, the Transformer architecture has gained attention in deep learning for its ability to model long-range dependencies, e.g., ChatGPT, which makes it suitable for time series modeling \cite{2022transformersSurvey}. In this paper, we leverage the high-performing Frequency Enhanced Decomposed Transformer (FEDformer) \cite{2022FEDformer} to model the auto-correlation of wind speed series. However, we observe that FEDformer primarily operates on frequency domain information obtained through Fourier transformation and lacks traditional processing of time-domain information. To address this limitation, we introduce a multi-head self-attention mechanism that allows us to effectively handle time-domain information. By incorporating this mechanism, we enhance the model's ability to capture temporal features and improve its forecasting performance.


\indent In this study, we analyzed wind speed data from 25 wind farms to determine their correlation. We found that the spatial distribution of wind farms varies, resulting in potential time lags in the correlation of wind speeds. Additionally, wind speeds can vary with different wind directions. To address these challenges, we propose a dynamic complex adjacency matrix that can simultaneously consider the spatial correlation and time lags between wind speeds from different wind farms. Furthermore, it can reflect real-time changes in the correlation between wind speeds, capturing their evolving relationship.

\indent To capture the spatio-temporal features among neighboring wind farms, we integrate GAT and FEDformer models and incorporate a multi-head self-attention mechanism to effectively process time-domain information. As a result, we propose the GFST-WSF model.

 The main contributions of this paper are summarized as follows:\par
\hangafter 1
\hangindent=8.95 mm
\indent
(a)  We propose a complex adjacency matrix to represent the spatio-temporal correlation among neighboring wind farms. Each element of the matrix is a complex number, where the real part represents the strength of the correlation, and the imaginary part represents the time difference of the correlation's occurrence. This approach provides a more comprehensive representation of the spatio-temporal correlation, leading to better forecasting performance in the proposed GFST-WSF model.\par
\hangafter 1
\hangindent=8.95 mm
\indent
(b) We enhance the GFST-WSF model with the GAT to effectively utilize the complex adjacency matrix and capture intricate spatial dependencies among neighboring wind farms’ wind speed. This approach improves the model's ability to extract spatial features, which is crucial for wind speed forecasting accuracy.\par
\hangafter 1
\hangindent=8.95 mm
\indent
(c) We introduce the FEDformer architecture, which leverages a frequency-enhanced mechanism to improve the temporal information extraction process in wind speed forecasting. We have further improved the FEDformer by adding the multi-headed attention mechanism to enhance its ability to capture long and short-term dependencies in the wind speed series. The deep architecture and multi-head attention of the GFST-WSF model enables it to effectively extract the temporal correlation within the wind speed from neighboring wind farms.

The rest of this paper is organized as follows: Section II presents the problem description; Section III presents the framework of GFST-WSF; Section IV presents the case studies; conclusions are drawn in Section V.

\section{Problem Description}

\indent We focus on the problem of short-term spatial-temporal wind speed forecasting for multiple wind farms. The meteorological data (including wind speed, wind direction, air pressure, temperature, etc.) of the $i$-th wind farm at time period $t$ can be denoted as $x_t^i$. 
 Thus, the data vector of all the wind farms at time $t$ is $\mathbf{x}_t=\{x_t^1, x_t^2,x_t^3,...,x_t^n \}$, where $n$ represents the total number of wind farms. We use multiple variables to forecast the wind velocity vector $p_{t+r}$ of a single wind farm, where $r\in Z^{+}$ represents the lead time.

Therefore, the forecasting model can be formulated as follows:
\begin{equation} \label{}
    f(\mathbf{x}_t,\mathbf{x}_{t-1},\mathbf{x}_{t-2},...\mathbf{x}_{t-(m-1)},\theta, \mathcal{G})=\hat{p}_{t+r}
\end{equation}
where $f(\cdot)$ represents mapping function, $\theta$ represents the parameters of $f(\cdot)$, $\mathcal{G}$ represents the spatial relationships among wind farms, $\hat{p}_{t+r}$ is the forecasted value of $p_{t+r}$, $m$ is the historical period for forecasting.

\begin{figure*}[t]
  \centering
  \includegraphics[width=0.9\linewidth]{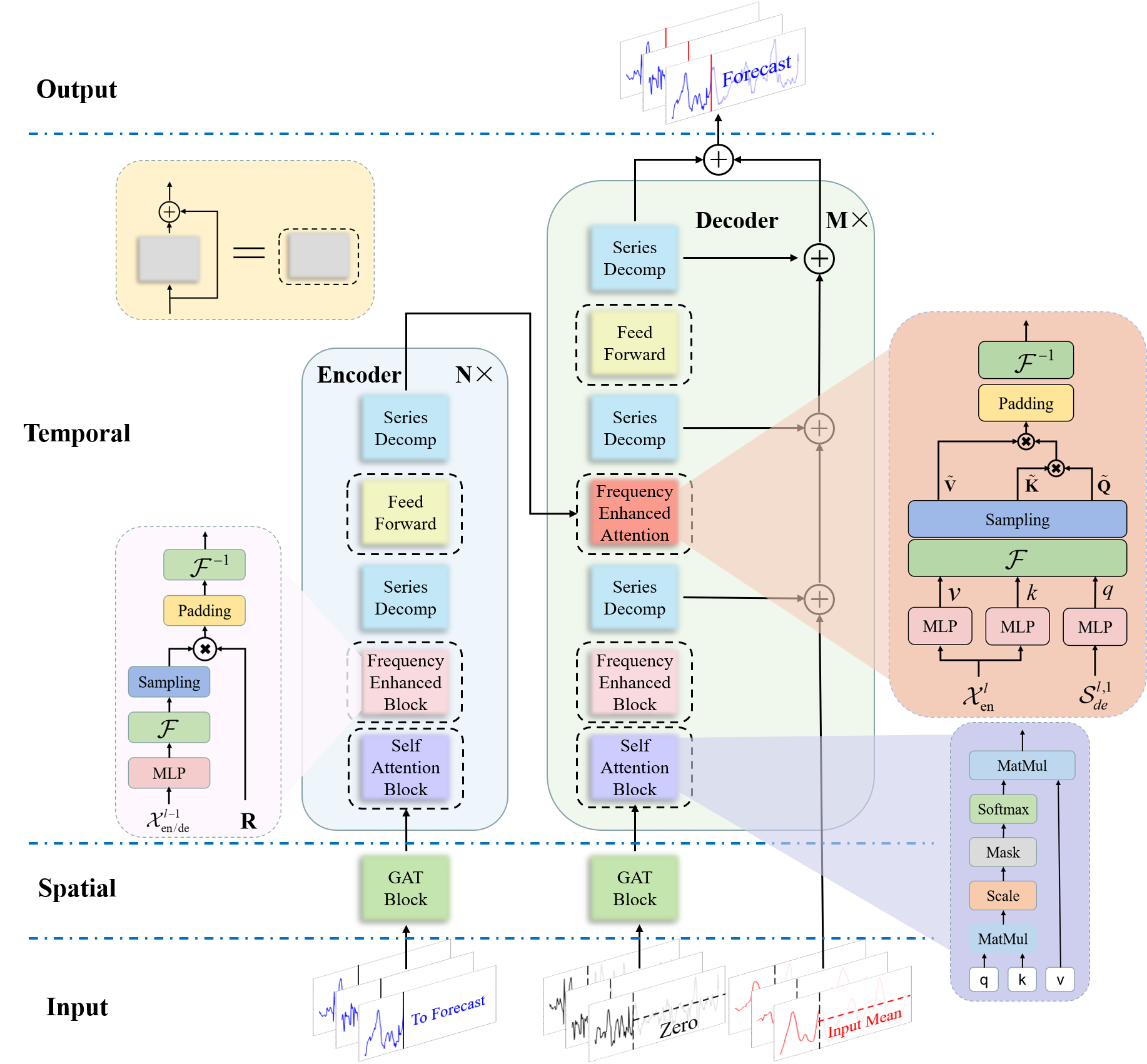}
  \caption{The GFST-WSF Structure. The Self-Attention Block utilizes the original multi-headed attention mechanism in Transformer. The GAT Block is composed of the time series transform module (TST) and the graph attention network (GAT). The Frequency Enhanced Block (FEB) and Frequency Enhanced Attention (FEA) are utilized to perform representation learning in frequency domain. The series decomposition blocks (SeriesDecomp) are employed to decompose the series into trend-cyclical and seasonal parts.}
  \label{fig:model}
  \vspace{-1em}
\end{figure*}

\section{The Framework of GFST-WSF}

\indent Inspired by recent innovations in the field of time series forecasting, GFST-WSF adopts an encoder-decoder structure, which includes graph neural network encoding, multi-head attention from the original Transformer, sequence decomposition using average pooling, frequency-enhanced mechanism from FEDformer for sequence-level connections, and feedforward networks, as shown in Fig.\ref{fig:model}.

\begin{figure*}[t]
  \centering
  \includegraphics[width=1.0\linewidth]{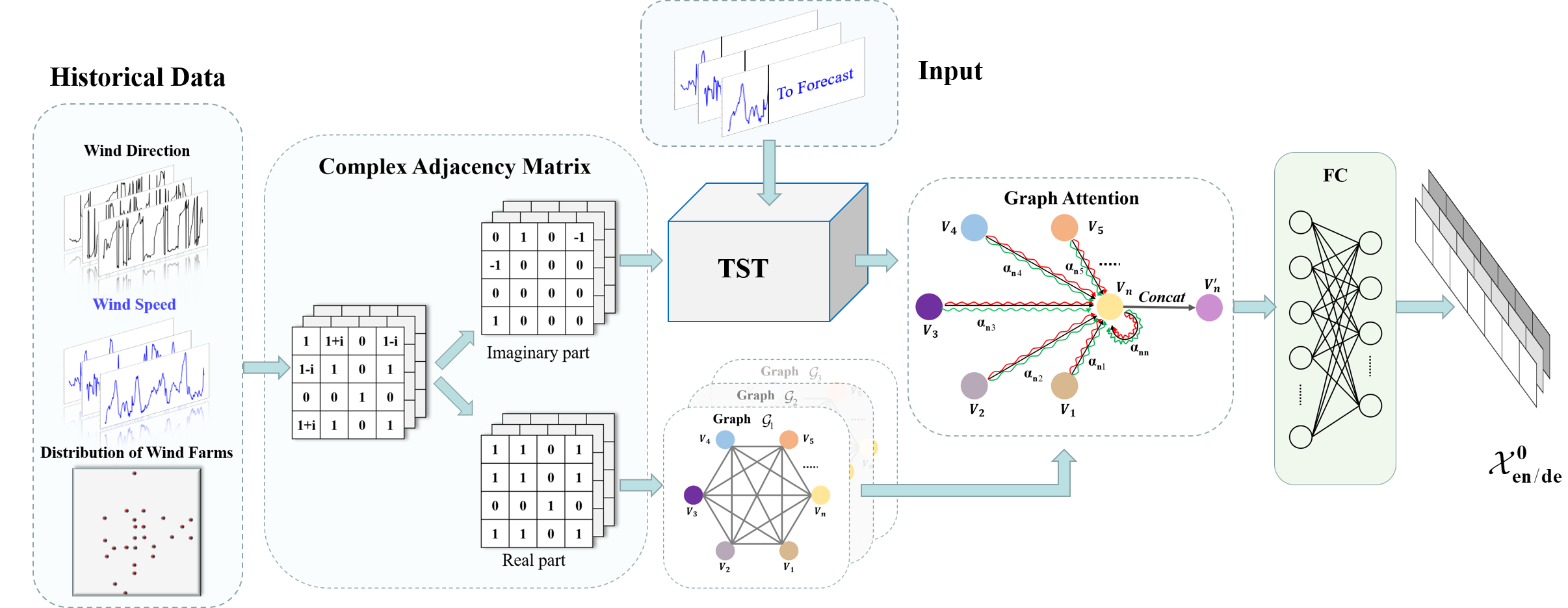}
  \caption{GAT Block. Constructing a complex adjacency matrix based on historical data (illustrated by a simple example in the figure), where the upper part represents the real components and the lower part represents the imaginary components. }
  \label{fig:GAT Block}
  \vspace{-1em}
\end{figure*}

\subsection{GAT Block}

\indent The GAT Block, illustrated in Fig.\ref{fig:GAT Block}, consists of two components: a GAT and a Time Series Transform module (TST). Initially, the dynamic complex adjacency matrix $\Lambda$ is decomposed into a real part matrix $\mathcal{A}$ and an imaginary part matrix $\mathcal{B}$. The elements of $\mathcal{A}$ reflect the correlation between wind farms, while the elements of $\mathcal{B}$ represent the time difference between them. In the TST module, the original time series features are transformed according to the elements of $\mathcal{B}$. Then, the updated time series features are used as input to the GAT, with the adjacency matrix of GAT being $\mathcal{A}$. Finally, the learned spatio-temporal information is utilized as input to the encoder.

\subsubsection{Building a Dynamic Complex Adjacency Matrix} 

\indent In this study, we propose a novel approach to represent the correlation between all wind farms in a given graph. Traditionally, the correlation between wind farms is represented using a simple adjacency matrix, where the nodes are the wind farms and the edges represent the correlation between them. However, due to the variability of wind speed, distance between wind farms, and constantly changing wind direction, the correlation between adjacent wind farms may exist at different time lags. To address this issue, we introduce a dynamic complex adjacency matrix denoted as $\Lambda$ to represent the time difference in the correlation between wind farms. This matrix contains complex values of the form $a+bi$, as illustrated in Fig.\ref{fig:GAT Block}. The real part of the complex value, $a$, represents the correlation between wind farms. If $a>0$, it indicates a positive correlation between wind farms. On the other hand, the imaginary part of the complex value, $b$, represents the time difference in the correlation between wind farms. If there is no correlation between wind farms, the value of $b$ is invalid. The complex adjacency matrix is defined as follows:
\begin{equation}
\Lambda(s_i,s_j)=\begin{cases} a+bi& \text{if}\ \ a>0 \\ 0&\text{otherwise} \end{cases}
\end{equation}
And the definition of the parameter $a$ is given as follows:
\begin{equation}\resizebox{0.5\hsize}{!}{$
\begin{aligned}
    a=\begin{cases} 1& if \ \ \ F(s_i,s_j)>\beta \\ 0&otherwise \end{cases}
    \end{aligned}
$}
\end{equation}
where $\beta$ is the threshold of the correlation coefficient, $s_i$ and $s_j$ represent the wind speed series of the $i$-th and $j$-th wind farm, and $F(\cdot)$ is a function that calculates the correlation between the two wind farms. In our study, we use the Pearson correlation coefficient, which is defined as follows:

\begin{equation}
F(s_i,s_j)=\frac{\sum{(s_i-\bar{s}_i)(s_j-\bar{s}_j)}}{\sqrt{\sum{(s_i-\bar{s}_i)^2} \sum{(s_j-\bar{s}_j)^2}}}
\end{equation}

To calculate the time difference $b$ between any two wind farms, we use the latitude and longitude information in the dataset to determine the azimuth information. We also apply the Haversine formula \cite{1957cosine_Haversine} to calculate the distance between any two points. Based on these values, we factor in the azimuth, wind direction, and distance to determine the time difference $b$.

\subsubsection{Graph Attention Networks} 

\indent Wind is a rapidly propagating air flow that affects a large area, resulting in intrinsic correlations between the wind speeds of adjacent wind farms. To capture these relationships, we model each wind farm as a graph node and the meteorological information series for the wind farm as the node's characteristic quantity. The correlation between wind farms is captured as the attribute of the graph's edges.

\indent  GNN \cite{2020GNN_review} refers to a general class of models that apply neural networks to graphs. These models can be categorized into different types based on their underlying techniques. However, as wind direction changes over time, the correlation between adjacent wind farms can also change. To account for this dynamic relationship, we employ GAT \cite{2017GAT}, which leverage masked self-attentional layers to overcome the limitations of prior methods, such as GCN and Graph Sample and Aggregate (GraphSAGE), that cannot handle dynamic graph relationships. 

\indent The input of GAT is a set of node features, $\mathbf{h}=\{\vec{h}_1,\vec{h}_2,...,\vec{h}_n  \}$, where $n$ is the number of nodes. The importance of node $j$'s features to node $i$ is defined as:
\begin{equation}\resizebox{0.4\hsize}{!}{$
\begin{aligned}
    e_{ij}=a(\mathbf{W}\vec{h}_i,\mathbf{W}\vec{h}_j)
    \end{aligned}
$}
\end{equation}
where $\mathbf{W}$ is a weight matrix, and the attention mechanism is a single-layer feedforward neural network. Meanwhile, use the softmax function to normalize them:
\begin{equation}\resizebox{0.7\hsize}{!}{$
\begin{aligned}
    \alpha_{ij}=softmax(e_{ij})=\frac{exp(e_{ij})}{\sum_{k\in \mathcal{N}_i}exp(e_{ik})}
    \end{aligned}
$}
\end{equation}

\indent Furthermore, GAT introduces the idea of multi-headed attention mechanism into graph neural networks. Specifically, perform $k$ separate attention calculations and then concatenate them:
\begin{equation}\resizebox{0.6\hsize}{!}{$
\begin{aligned}
    \vec{h}_{i}^{\prime}=\|_{k=1}^{K} \sigma\left(\sum_{j \in \mathcal{N}_{i}} \alpha_{i j}^{k} \omega^{k} \vec{h}_{j}\right)
    \end{aligned}
$}
\end{equation}
where $\|$ is the concatenation operation and $\omega^k$ is the corresponding input linear transformation's weight matrix.

\subsection{Encoder}
\indent The encoder comprises multiple layers, with the output of the $l$-th layer denoted as $\mathcal{X}_{en}^l=Encoder(\mathcal{X}_{en}^{l-1})$, where $l \in \{1,2,..,N\}$. The input to the encoder is $\mathcal{X}_{en}^0$, which is the historical series that has been embedded using a graph attention network and one-dimensional convolution. The spatio-temporal information learned from this embedding is fed to the encoder to enable learning of the sequence through frequency enhanced blocks, which are then decomposed using average pooling. To capture interdependent features in the sequence and increase model variability, multi-head attention is used. The overall learning process is formulated as follows:
\begin{align}
    \mathcal{X}_{en}^0 &= Conv1D(GAT(\mathcal{X},\mathcal{G}))\\
    \mathcal{I}_{en}^{l} &= Attention(\mathcal{X}_{en}^{l-1})+\mathcal{X}_{en}^{l-1}\\
    \mathcal{S}_{en}^{l,1},\_ &= SeriesDecomp(FEB(\mathcal{I}_{en}^{l})+\mathcal{I}_{en}^{l})\\
    \mathcal{S}_{en}^{l,2},\_ &= SeriesDecomp(FeedForward(\mathcal{S}_{en}^{l,1})+\mathcal{S}_{en}^{l,1})\\
    \mathcal{X}_{en}^l &= \mathcal{S}_{en}^{l,2}
\end{align}

\noindent where $\mathcal{X}$ represents the meteorological information of all wind farms, $\mathcal{G}$ represents the graph network code of the sites, $\mathcal{I}_{en}^{l}$ is the result of multi-head attention, and $\mathcal{S}_{en}^{l,i}, i\in {1,2}$ represents the seasonal component after the $i$-th series decomposition block in the $l$-th layer respectively. The frequency enhanced block (FEB) module is implemented using a Discrete Fourier Transform (DFT) mechanism.



\subsubsection {Frequency Enhanced Block with Fourier Transform (FEB)} 

\indent The structure presented in this paper leverages the Discrete Fourier Transform (DFT), which was first introduced in \cite{2022FEDformer}. By employing the Fast Fourier Transform (FFT) and randomly selecting a fixed number of Fourier components, the computational complexity of the structure can be reduced to $O(N)$, making it more efficient.

\indent The FEB block takes an input $\boldsymbol{x}\in \mathbb{R}^{N\times D}$, which is first linearly projected using $\boldsymbol{w} \in \mathbb{R}^{D \times D}$, and then converted to the frequency domain using the Fourier transform. Specifically, the input is transformed as follows: $\boldsymbol{q}=\boldsymbol{x}\cdot \boldsymbol{w},\ \ \boldsymbol{Q}=\mathcal{F}(\boldsymbol{q})$, where $\mathcal{F}$ represents the Fourier transform, and $\boldsymbol{Q} \in \mathbb{C}^{N \times D}$. In the frequency domain, only a randomly selected subset of $M(M<<N)$ modes are retained, resulting in $\boldsymbol{\widetilde{Q}}=Select(\boldsymbol{Q})$, where $\boldsymbol{\widetilde{Q}} \in \mathbb{C}^{M \times D}$. Thus, the FEB is defined as
\begin{equation}\resizebox{0.6\hsize}{!}{$
\begin{aligned}
    FEB(\boldsymbol{q})=\mathcal{F}^{-1}(Padding(\widetilde{\boldsymbol{Q}}\odot \boldsymbol{R}))
    \end{aligned}
$}
\end{equation}
where $\boldsymbol{R} \in \mathbb{C}^{D \times D \times M}$ is a parameterized kernel initialized randomly, $\mathcal{F}^{-1}$ denotes the inverse Fourier transform. The operator $\odot$ is the complex matrix multiplication. I.e., the result is zero-padded to $\mathbb{C}^{N \times D}$ and converted back to the time dimension by $\mathcal{F}^{-1}$.

\subsubsection {Series Decomp} 

\indent In order to model the complex temporal patterns in wind speed sequences, the decomposition approach is adopted to divide the time series into two components: the trend-cyclical component and the seasonal component, which respectively reflect the long-term progress and periodicity of the time series. In practice, the moving average method is used to smooth out the cyclical fluctuations and highlight the long-term trend. Specifically, for an input sequence $\mathcal{X}$ of length $L$, the decomposition process is as follows:
\begin{align}
    \mathcal{X}_{t} &= AvgPool(Padding(\mathcal{X}))\\
    \mathcal{X}_{s} &= \mathcal{X} - \mathcal{X}_{t}
\end{align}
\noindent where $\mathcal{X}_{t}, \mathcal{X}_{s}$ denote the trend-cyclical and the seasonal part respectively. We use the $AvgPool(·)$ operation with padding for moving average and summarize the above equations using $\mathcal{X}_{t}, \mathcal{X}_{s} = SeriesDecomp(\mathcal{X})$.



\subsection{Decoder}
\indent The decoder also adopts a multilayer structure as : $\mathcal{X}_{de}^l,\mathcal{T}_{de}^{l}=Decoder(\mathcal{X}_{de}^{l-1},\mathcal{T}_{de}^{l-1})$, where $l \in \{ 1,2,...,M\}$ is the output of $l$-th decoder layer. The decoder takes two inputs: $\mathcal{X}_{de}$, which is the spatio-temporal embedding obtained from the seasonal initialization term through graph attention and one-dimensional convolution, and $\mathcal{T}_{de}$, which is the direct input of the trend initialization term. The first layer of the decoder, $\mathcal{X}_{de}^0$, is obtained by applying a one-dimensional convolution to the output of a graph attention layer that takes as input the concatenation of the average-pooled $\mathcal{X}$ and the seasonal initialization term $\mathcal{S}_{init}$, with attention weights given by the graph $\mathcal{G}$. In each subsequent layer $l$, the input $\mathcal{X}_{de}^{l-1}$ is refined by an attention mechanism and added to the output of the encoder for the following decoding. The trend component $\mathcal{T}_{de}$ is accumulated across the layers to improve the model's inference ability. Specifically, we define the decomposition blocks as follows:
\begin{align}
    \mathcal{S}_{de}^{0} &= Concat(SeriesDecomp(\mathcal{X}), \mathcal{S}_{init})\\
    \mathcal{X}_{de}^0 &= Conv1D(GAT(\mathcal{S}_{de}^{0},\mathcal{G}))\\
    \mathcal{I}_{de}^{l} &= Attention(\mathcal{X}_{de}^{l-1})+\mathcal{X}_{de}^{l-1}\\
    \mathcal{T}_{de}^0 &= Concat(SeriesDecomp(\mathcal{X}),\mathcal{T}_{init})\\
    \mathcal{S}_{de}^{l,1},\mathcal{T}_{de}^{l,1} &= SeriesDecomp(FEB(\mathcal{I}_{de}^{l})+\mathcal{I}_{de}^{l})\\
    \mathcal{S}_{de}^{l,2},\mathcal{T}_{de}^{l,2} &= SeriesDecomp(FEA(\mathcal{S}_{de}^{l,1},\mathcal{X}_{en}^{N})+\mathcal{S}_{de}^{l,1})\\
    \mathcal{S}_{de}^{l,3},\mathcal{T}_{de}^{l,3} &= SeriesDecomp(FeedForward(\mathcal{S}_{de}^{l,2})+\mathcal{S}_{de}^{l,2})\\
    \mathcal{T}_{de}^{l} &= \mathcal{T}_{de}^{l-1} + \mathcal{W}_{l,1}\cdot \mathcal{T}_{de}^{l,1}+ \mathcal{W}_{l,2}\cdot \mathcal{T}_{de}^{l,2}+ \mathcal{W}_{l,3}\cdot \mathcal{T}_{de}^{l,3}\\
    \mathcal{X}_{de}^l &= \mathcal{T}_{de}^{l,3}
\end{align}

\noindent where $\mathcal{S}_{init}$ and $\mathcal{T}_{init}$ represent the initialization of seasonal and trend terms respectively. After the $i$-th series decomposition block in the $l$-th layer, the seasonal and trend components are represented by $\mathcal{S}_{de}^{l,i}$ and $\mathcal{T}_{de}^{l,i}$, where $i \in \{1,2,3\}$. The linear projector for the $i$-th extracted trend $\mathcal{T}_{de}^{l,i}$ is denoted by $\mathcal{W}_{l,i}$. The frequency enhanced attention (FEA) is similar to FEB in that it uses DFT projection with an attention design. 

\ 

\noindent
\textbf{Frequency Enhanced Attention with Fourier Transform(FEA)} FEA takes inputs that are similar to those of the original Transformer, including queries  $\boldsymbol{q} \in \mathbb{R}^{L \times D}$, keys $\boldsymbol{k} \in \mathbb{R}^{L \times D}$, and values $\boldsymbol{v} \in \mathbb{R}^{L \times D}$. The queries are sourced from the decoder, while the keys and values come from the encoder : $ \boldsymbol{q}=\boldsymbol{x}_{de} \cdot \boldsymbol{w}_q, \ \boldsymbol{k}=\boldsymbol{x}_{en} \cdot \boldsymbol{w}_k, \ \boldsymbol{v}=\boldsymbol{x}_{en} \cdot \boldsymbol{w}_v $
, where $\boldsymbol{w}_q,\boldsymbol{w}_k,\boldsymbol{w}_v \in \mathbb{R}^{D \times D}$. Then the canonical attention can be defined as 
\begin{equation}\resizebox{0.6\hsize}{!}{$
\begin{aligned}
    Atten(\boldsymbol{q},\boldsymbol{k},\boldsymbol{v})=Softmax(\frac{\boldsymbol{q}\boldsymbol{k}^{T}}{\sqrt{d_q}})\boldsymbol{v}
    \end{aligned}
$}
\end{equation}
where $d_q$ represents the length of $\boldsymbol{q}$. In FEA, the queries, keys, and values ($\boldsymbol{q}, \boldsymbol{k}$, and $\boldsymbol{v}$) are first transformed into the frequency domain using the Fourier transform, after which a subset of $M$ modes is randomly selected in the frequency domain. Next, a similar cross-attention mechanism is applied in the frequency domain. The FEA can be defined as follows:
\begin{align}
    FEA(\boldsymbol{q},\boldsymbol{k},\boldsymbol{v}) &= \mathcal{F}^{-1}(Padding(\sigma(\widetilde{\boldsymbol{Q}} \cdot \widetilde{\boldsymbol{K}}) \cdot \widetilde{\boldsymbol{V}}))
\end{align}

where $\sigma$ is the activation function, $\widetilde{\boldsymbol{Q}} =Select(\mathcal{F}(\boldsymbol{q}))$, $\widetilde{\boldsymbol{K}} =Select(\mathcal{F}(\boldsymbol{k})),$ and $\widetilde{\boldsymbol{V}} =Select(\mathcal{F}(\boldsymbol{v}))$. The result of $\sigma(\widetilde{\boldsymbol{Q}} \cdot \widetilde{\boldsymbol{K}}) \cdot \widetilde{\boldsymbol{V}}$ also need to be zero-padded to $\mathbb{C}^{L \times D}$ before converting back to the time dimension.

\section{Case Studies}

\subsection{Evaluation Metrics}

\indent The Mean Squared Error (MSE) and normalized Mean Absolute Error (MAE) are commonly used as evaluation metrics in the task of forecasting wind speed at a single site. The metrics are defined as follows:
\begin{align}
    &MSE=\frac{1}{T}\sum_{t=1}^{T}(p_{t+r} - \hat{p}_{t+r})^2\\
    &MAE=\frac{1}{T}\sum_{t=1}^T|p_{t+r} - \hat{p}_{t+r}|
\end{align}

\begin{figure}[t!] 
  \centering
  \includegraphics[width=3in]{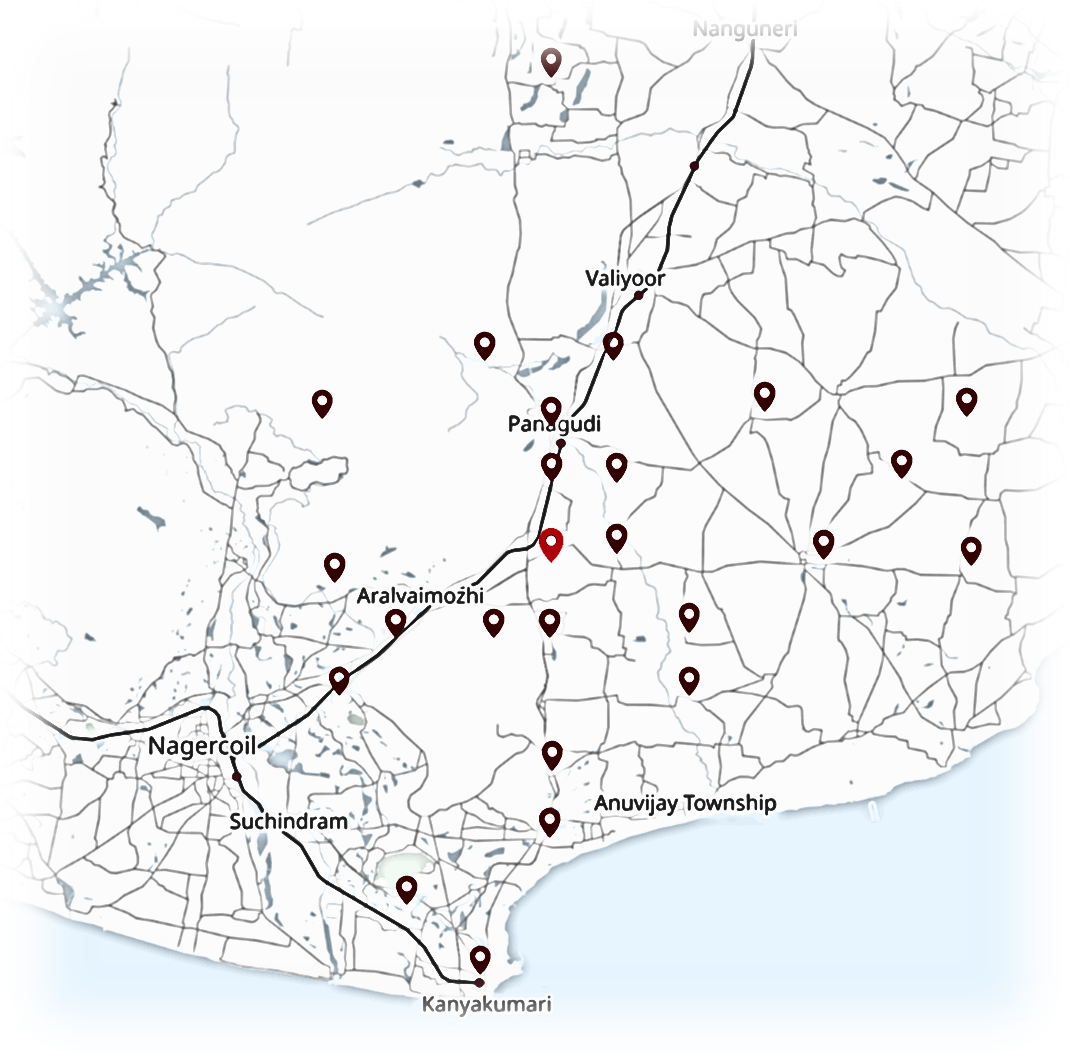}
  \caption{The geographic distribution of the wind farms. These wind farms are located in southern India and have been clearly marked on the map.}
  \label{fig:map}
\end{figure}

\subsection{Data Description}
\indent The Wind Integration National Dataset (WIND) Toolkit \cite{2015windtoolkit} is a software package developed by the National Renewable Energy Laboratory (NREL) in the United States. It provides researchers with access to a wide range of wind power data. For this study, we utilized the India Wind Dataset from WIND, which was developed in part through the India Renewable Integration Study. This dataset contains wind speed, wind direction, temperature, and pressure at heights of 40m, 80m, 100m, and 120m above the ground. The data has a spatial resolution of 3 km and a temporal resolution of 5 minutes. From this dataset, we selected 25 sites in Tamil Nadu, India for short-term wind speed forecasting, as shown in Fig.\ref{fig:map}. The annual dataset for 2014 was chosen and the time resolution was adjusted to 15 minutes. The wind farm being forecasted is located at the center of the selected sites and is the largest wind farm in southern India, named the Muppandal Wind Farm (the green point). The distance between sites ranges from 5 kilometers to 40 kilometers, as shown in the Fig.\ref{fig:map}.

\subsection{Comparison on Test Set} To verify the superiority of the GFST-WSF, we employed state-of-the-art time series forecasting methods and deep learning methods as baselines. There are following: \par
\hangafter 1
\hangindent=8.95 mm
\indent 
1) The Persistence method (Pers.): The Persistence method forecasts the next time step by taking the current time step's observed value as the forecasted value, i.e., all forecasted values are equal to the last observed value at the known time step.\par
\hangafter 1
\hangindent=8.95 mm
\indent 
2) LightGBM: LightGBM is a mainstream implementation of the Gradient Boosting Decision Tree (GBDT) algorithm, which utilizes iteratively trained weak classifiers (decision trees) to obtain an optimized model. It is commonly used for forecasting tasks.\par
\hangafter 1
\hangindent=8.95 mm
\indent 
3) Long Short-Term Memory (LSTM): LSTM, as a deep learning algorithm used for processing sequence data, features memory cells and gate mechanisms, exhibiting strong abilities in modeling long-term dependencies and effectively addressing the issues of vanishing and exploding gradients.\par
\hangafter 1
\hangindent=8.95 mm
\indent 
4) Transformer: Transformer is a neural network architecture based on self-attention mechanism used for sequence-to-sequence tasks and is one of the state-of-the-art models in this field. \par
\hangafter 1
\hangindent=8.95 mm
\indent 
5) FEDformer: FEDformer is a variant of the Transformer model that introduces frequency domain enhancement mechanisms, and currently demonstrates excellent performance in time series forecasting tasks.

\begin{figure*}[t]
\centering
\captionsetup[subfloat]{labelsep=space, margin={15pt,10pt}}
\subfloat[Results on test set (6-hour ahead)]{\includegraphics[width=2in]{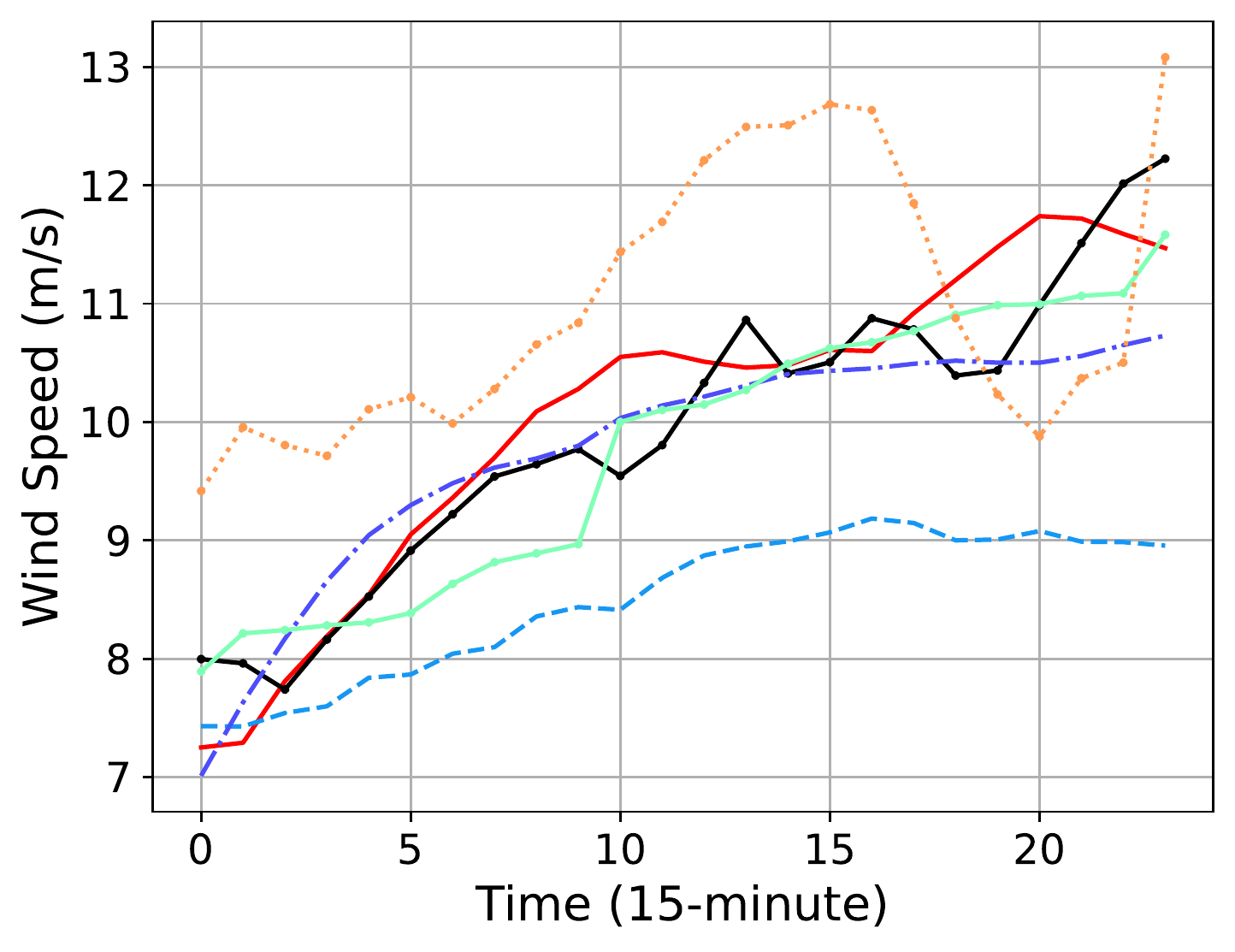}
\label{24_baselines}}
\captionsetup[subfloat]{labelsep=space, margin={15pt,10pt}}
\subfloat[Results on test set (12-hour ahead)]{\includegraphics[width=2in]{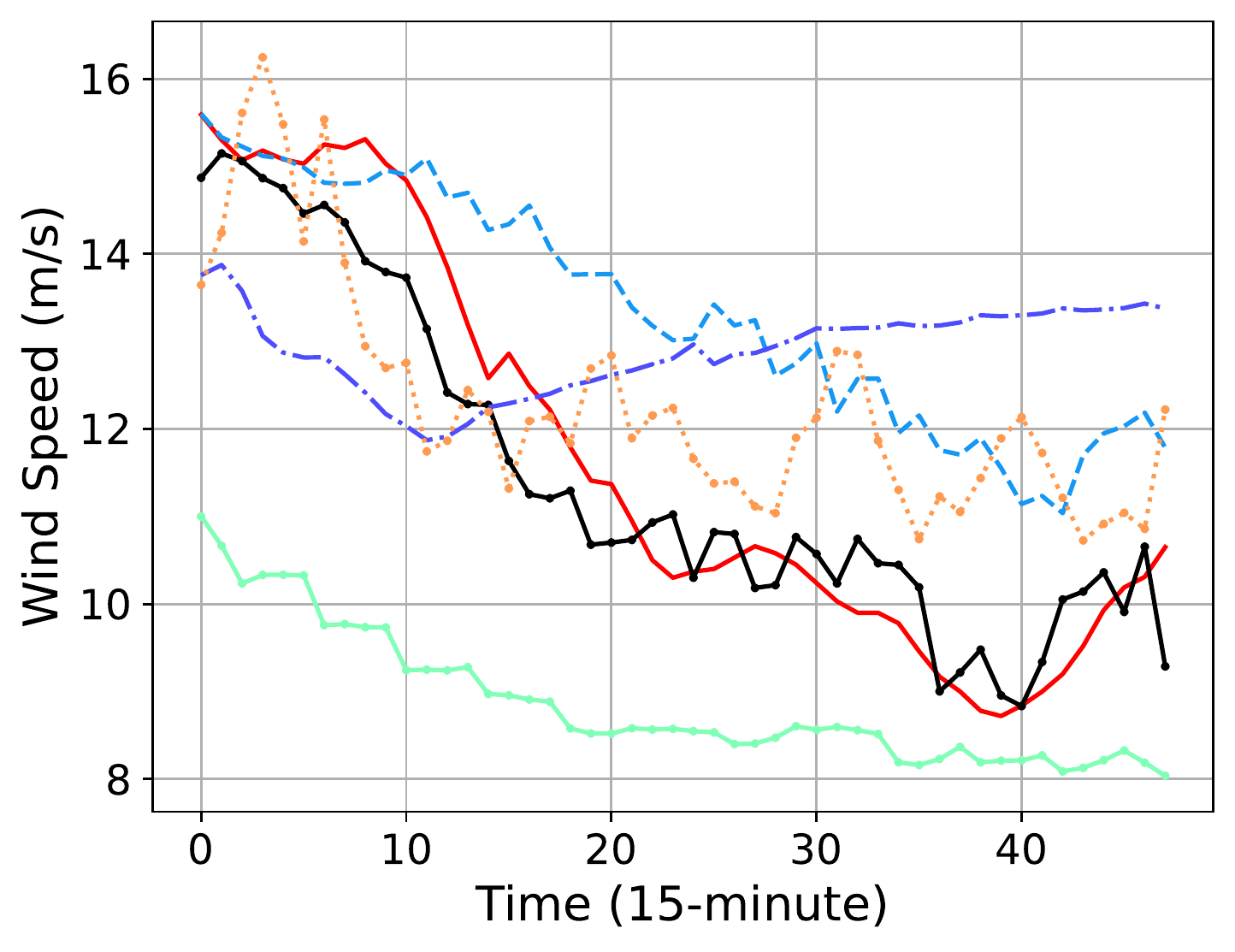}
\label{48_baselines}}
\captionsetup[subfloat]{labelsep=space, margin={10pt,60pt}}
\subfloat[Results on test set (24-hour ahead)]{\includegraphics[width=2.8in]{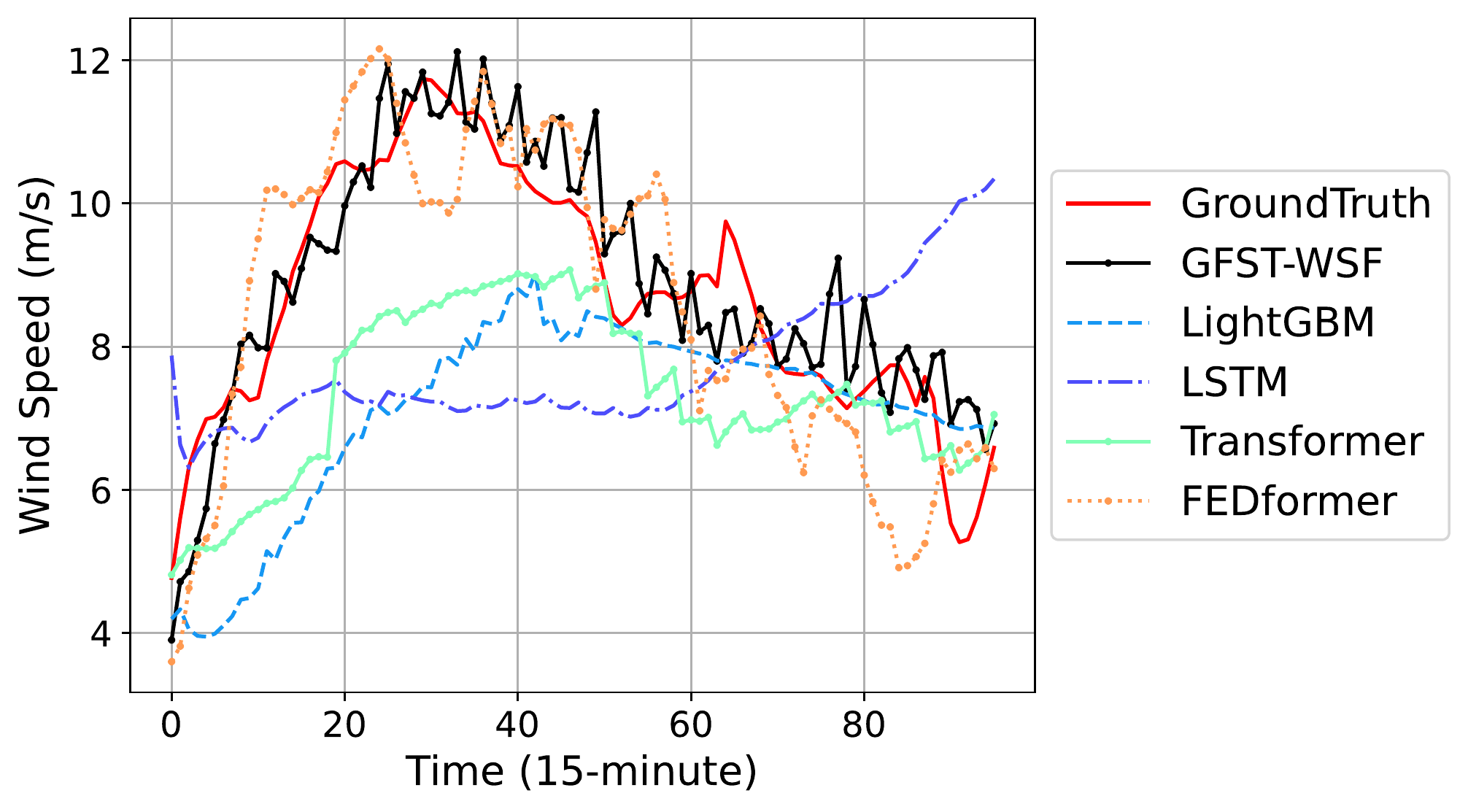}
\label{96_baselines}}
\vspace{-0.0cm}
\caption{Comparison between GFST-WSF and baselines.}
\label{PF1}
\vspace{-1em}
\end{figure*}

\begin{figure*}[t]
\centering
\captionsetup[subfloat]{labelsep=space, margin={15pt,10pt}}
\subfloat[Results on test set (6-hour ahead)]
{\includegraphics[width=2in]{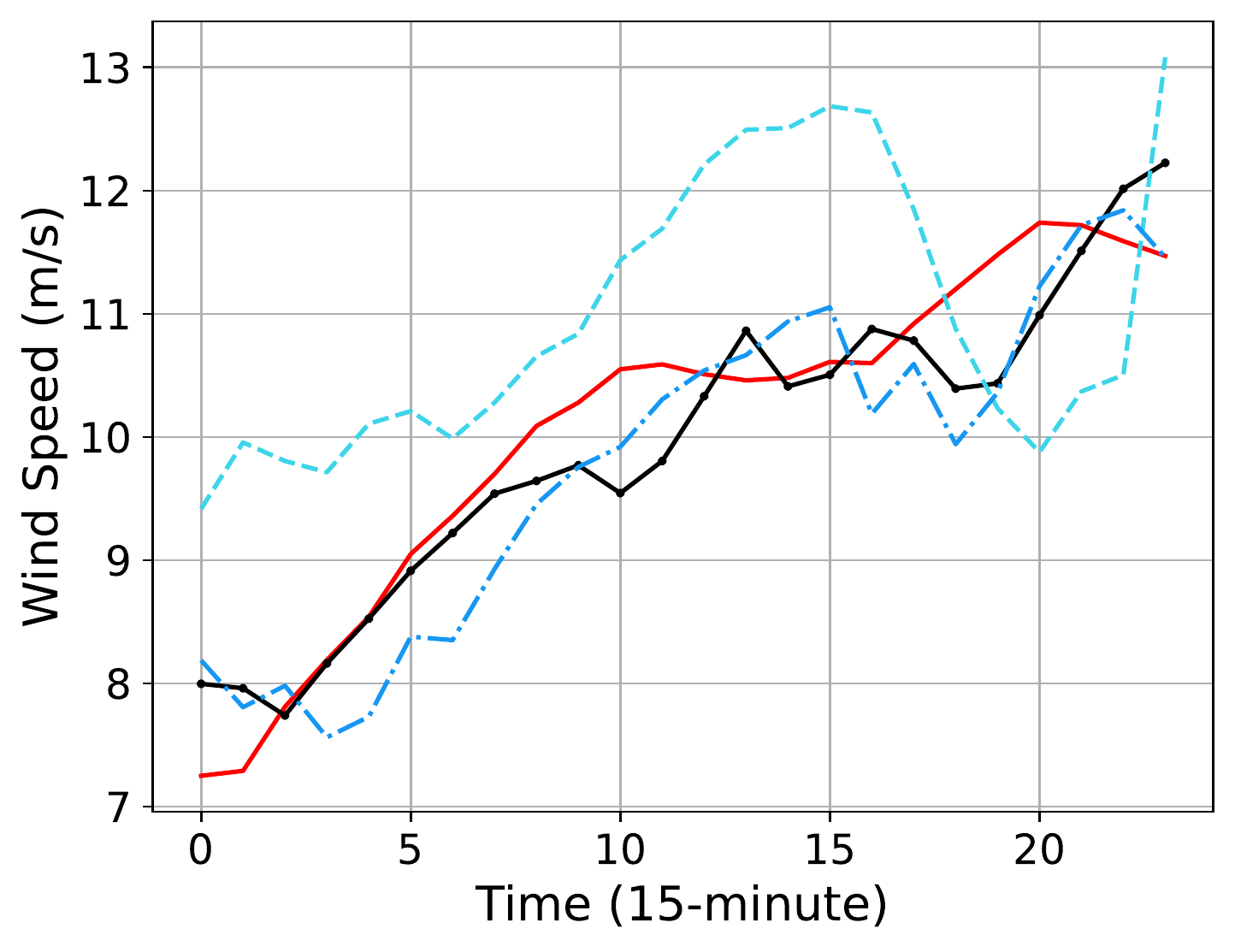}
\label{24_ablation}} 
\captionsetup[subfloat]{labelsep=space, margin={15pt,10pt}}
\subfloat[Results on test set (12-hour ahead)]
{\includegraphics[width=2in]{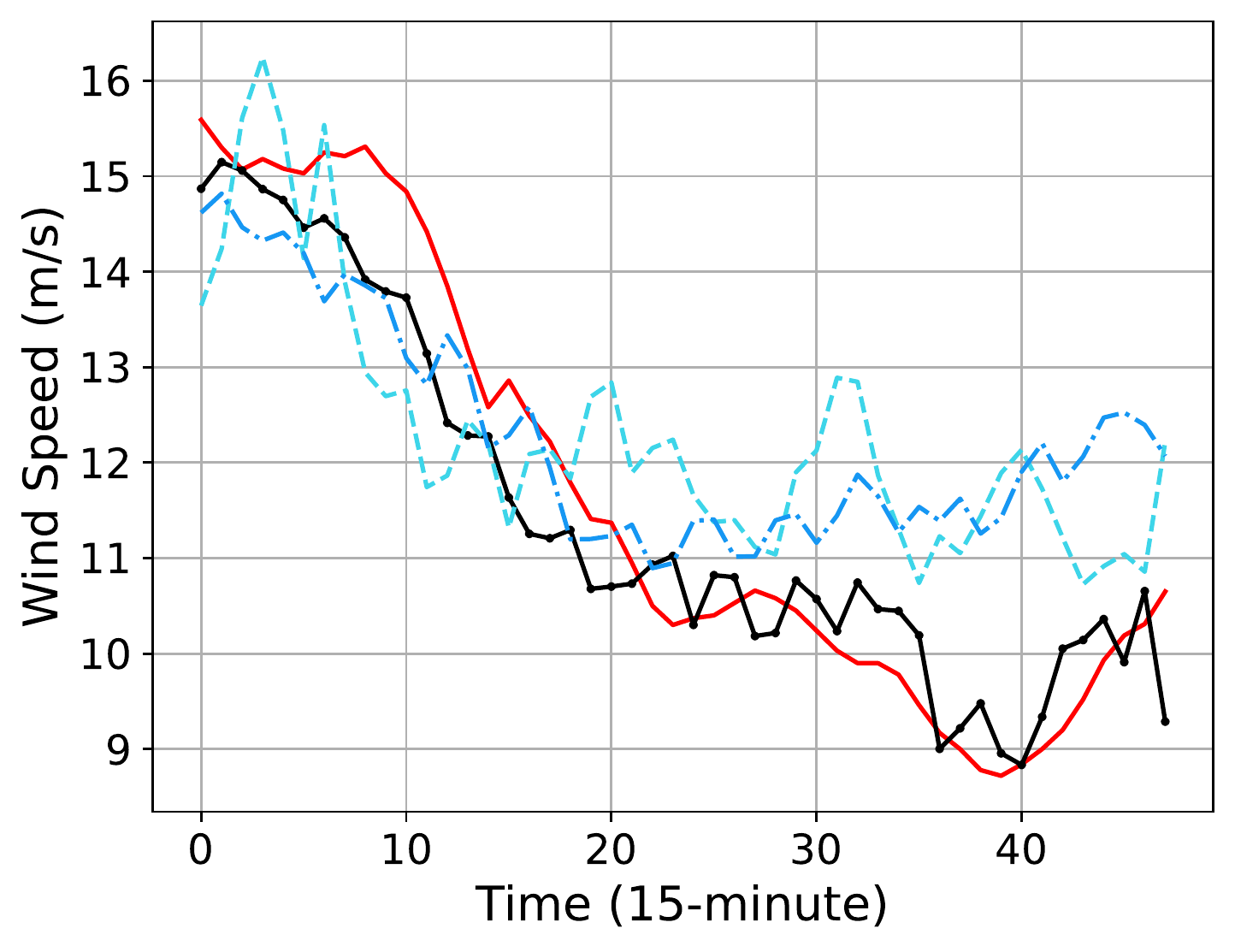}
\label{48_ablation}}
\captionsetup[subfloat]{labelsep=space, margin={10pt,60pt}}
\subfloat[Results on test set (24-hour ahead)]
{\includegraphics[width=2.9in]{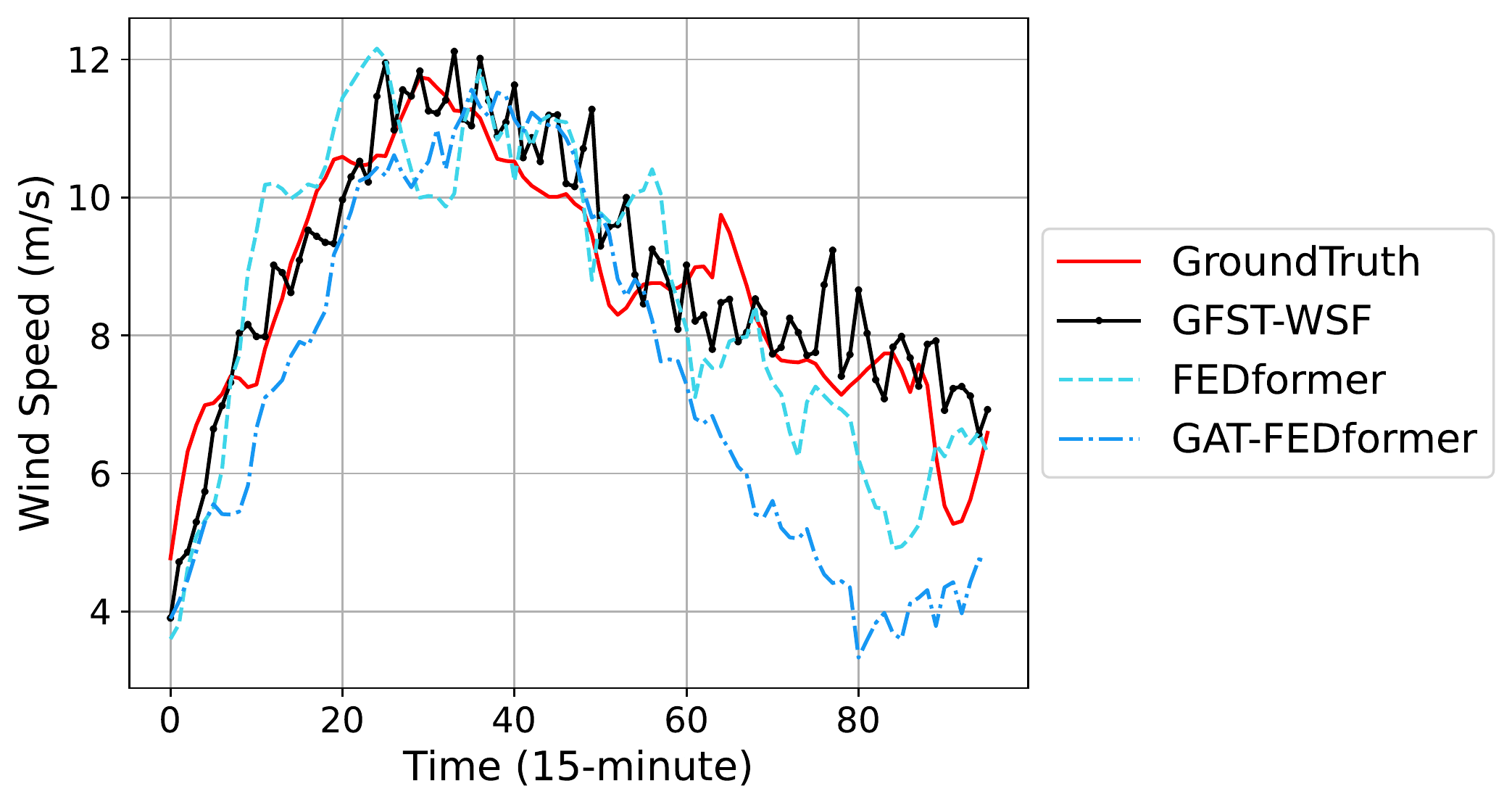}
\label{96_ablation}}
\vspace{-0.0cm}
\caption{Comparison of results from ablation experiments.}
\label{Ablation}
\vspace{-1em}
\end{figure*}

\begin{table}[t]
\caption{MSE of wind speed in the next 6h, 12h and 24h by different models.}
\vspace*{3mm}
\label{MSE}
\centering
\footnotesize
\begin{tabular}{c|c|c|c}
\hline\hline
\multirow{2}{*}{Models} & \multicolumn{3}{c}{Forecasting Horizon} \\
\cline{2-2}\cline{3-3}\cline{4-4}
 & {24(6-hour)} & {48(12-hour)} & {96(24-hour)}\\
\hline
Pers. & $0.215$ & $0.378$ & $0.415$ \\
LightGBM & $0.164$ & $0.243$ & $0.284$ \\
LSTM & $0.175$ & $0.272$ & $0.366$ \\
Transformer & $0.165$ & $0.236$ & $0.339$ \\
FEDformer & $0.165$ & $0.225$ & $0.244$ \\
GFST-WSF & $\mathbf{0.141}$ & $\mathbf{0.197}$ & $\mathbf{0.226}$ \\
\hline \hline
\end{tabular}
\leftline{}\\
\leftline{\hspace{0em}$\rm \textbf{*}$  All of the data has been normalized.}
\leftline{\hspace{0em}$\rm \textbf{*}$  The experimental results are the mean of multiple experiments.}
\vspace{0em}
\end{table}

\begin{table}[t!]
\caption{MAE of wind speed in the next 6h, 12h and 24h by different models.}
\vspace*{3mm}
\label{MAE}
\centering
\footnotesize
\begin{tabular}{c|c|c|c}
\hline\hline
\multirow{2}{*}{{Models}} & \multicolumn{3}{c}{{Forecasting Horizon}} \\
\cline{2-2}\cline{3-3}\cline{4-4}
 & {24(6-hour)} & {48(12-hour)} & {96(24-hour)}\\
\hline
Pers. & $0.337$ & $0.466$ & $0.499$ \\
LightGBM & $0.299$ & $0.379$ & $0.414$ \\
LSTM & $0.314$ & $0.405$ & $0.489$ \\
Transformer & $0.317$ & $0.393$ & $0.473$ \\
FEDformer & $0.315$ & $0.373$ & $0.390$ \\
GFST-WSF & $\mathbf{0.282}$ & $\mathbf{0.344}$ & $\mathbf{0.374}$ \\
\hline \hline
\end{tabular}
\leftline{}\\
\leftline{\hspace{0em}$\rm \textbf{*}$  All of the data has been normalized.}
\leftline{\hspace{0em}$\rm \textbf{*}$  The experimental results are the mean of multiple experiments.}
\vspace{0em}
\end{table}

\

\indent In TABLE \ref{MSE} and TABLE \ref{MAE}, we present a comparison of the performance of our GFST-WSF model with other time series forecasting models. The results clearly demonstrate that GFST-WSF outperforms all considered methods in terms of the lowest MAE and MSE values. As a commonly used machine learning method, LightGBM performs well in short-term forecasting but exhibits poor performance in long-term forecasting. Among the baselines, LSTM performed the worst in terms of forecasting performance, with its error increasing sharply as the length of the forecasting horizon increased. However, the Transformer model showed improved performance, owing to its effective capture of the attention mechanism on the long-term and short-term dependence of time series. Building upon the Transformer architecture, FEDformer achieved significant performance improvements in forecasting long-term time series. Specifically, in the 12-hour forecasting task, FEDformer lowered MSE and MAE by approximately 5\%, while in the 24-hour forecasting task, it demonstrated a 28\% reduction in MSE and a 17\% reduction in MAE. This is because FEDformer replaced the multi-head self-attention mechanism in Transformer with a frequency-enhanced mechanism, and applied cross-attention to the frequency-domain information. Additionally, FEDformer utilized a temporal decomposition module, which improved its performance in long sequence forecasting.

\indent GFST-WSF is a model based on FEDformer, which combines GAT module to extract spatial features and uses multi-head attention mechanism to extract temporal information, thus further improving the forecasting performance. At the same time, GFST-WSF uses dynamic complex adjacency matrix to model the time lag relationship between adjacent wind farms, in order to better capture the wind speed correlation. Through these improvements, GFST-WSF can effectively extract various wind speed features and capture complex patterns in wind speed data, thereby performing well in short-term wind speed forecasting tasks. Compared with other baseline models, GFST-WSF has stronger modeling and representation capabilities, and therefore has better application prospects in ensuring the safe and stable operation of wind power integrated grids.

\indent The experimental results on the test set are shown in Fig.\ref{PF1}, where Fig.\ref{24_baselines}, Fig.\ref{48_baselines}, and Fig.\ref{96_baselines} represent the experimental results of all comparative models in forecasting future 6, 12, and 24 hours, respectively. It can be clearly observed that GFST-WSF is the closest to the ground truth in the results of different forecasting lengths. Especially as the forecasting length increases, the performance advantage of GFST-WSF becomes more pronounced.

\subsection{Ablation experiments}
\indent To verify the effectiveness and necessity of each module in GFST-WSF, we also conducted an additional ablation experiment, and the results are listed in TABLE \ref{Ablation_table}. After removing the multi-head self-attention, the new model is called GAT-FEDformer; then remove the GAT Block, and the model becomes the original FEDformer. TABLE \ref{Ablation_table} shows the following.\par
\hangafter 1
\hangindent=8.95 mm
\indent 
1) FEDformer has shown remarkable performance in forecasting longer time series, with its frequency-enhanced mechanism playing a crucial role. \par
\hangafter 1
\hangindent=8.95 mm
\indent 
2) GAT-FEDformer outperforms FEDformer by 5\%-12\% in forecasting wind speed at different lengths, indicating the effectiveness of the GAT and the designed dynamic complex adjacency matrix in capturing spatial features. \par
\hangafter 1
\hangindent=8.95 mm
\indent 
3) GFST-WSF surpasses GAT-FEDformer in terms of forecasting accuracy, demonstrating that the multi-head self-attention on temporal information can enhance the model's representation ability and improve its overall performance.

\begin{table}
 \centering
 \caption{Performance Comparison of Ablation Experiments}
 \vspace*{-2mm}
 \label{Ablation_table}
 \footnotesize
\begin{tabular}{c|c|c|c|c}
\hline \hline
\multirow{2}{*}{Models} & \multirow{2}{*}{Metric}& \multicolumn{3}{c}{Lead step}\\
\cline{3-3}\cline{4-4}\cline{5-5}
 && {24(6h)} & {48(12h)} & {96(24h)}\\
 \hline
\multirow{2}{*}{Pers.} & $\text{MSE}$& $0.215$ & $0.378$ & $0.415$   \\ 
        &   $\text{MAE}$ & $0.337$ & $0.466$ & $0.499$ \\ 
\cline{1-1}\cline{2-2}\cline{3-3}\cline{4-4}\cline{5-5}
\multirow{2}{*}{FEDformer} & $\text{MSE}$& $0.165$ & $0.225$ & $0.244$ \\ 
        & $\text{MAE}$ & $0.315$ & $0.373$ & $0.390$ \\ 
\cline{1-1}\cline{2-2}\cline{3-3}\cline{4-4}\cline{5-5}
\multirow{2}{*}{GAT-FEDformer} &  $\text{MSE}$& $0.157$ & $0.198$ & $0.232$  \\ 
        &  $\text{MAE}$ & $0.299$ & $0.345$ & $0.375$  \\
\cline{1-1}\cline{2-2}\cline{3-3}\cline{4-4}\cline{5-5}
\multirow{2}{*}{GFST-WSF} & $\text{MSE}$& $\mathbf{0.141}$ & $\mathbf{0.197}$ & $\mathbf{0.226}$  \\ 
        &  $\text{MAE}$ & $\mathbf{0.282}$ & $\mathbf{0.344}$ & $\mathbf{0.374}$  \\
\hline\hline
\end{tabular}
\leftline{}\\
\vspace{-1em}
\end{table}

\indent Based on the results of the ablation experiment, which are shown in Fig.\ref{Ablation} and TABLE \ref{Ablation_table}, it is evident that the GFST-WSF model exhibits the highest forecast performance. Specifically, the results indicate that each module plays a significant role in the model's overall performance, and that the removal of any one module significantly impairs the forecast capabilities of the model. Thus, the GFST-WSF model stands out as a superior choice for this task, owing to its comprehensive architecture and the high degree of synergy among its constituent parts.

\begin{figure}[t]
\centering
\subfloat[Distribution of MSE for 100 instances.]{\includegraphics[width=3in]{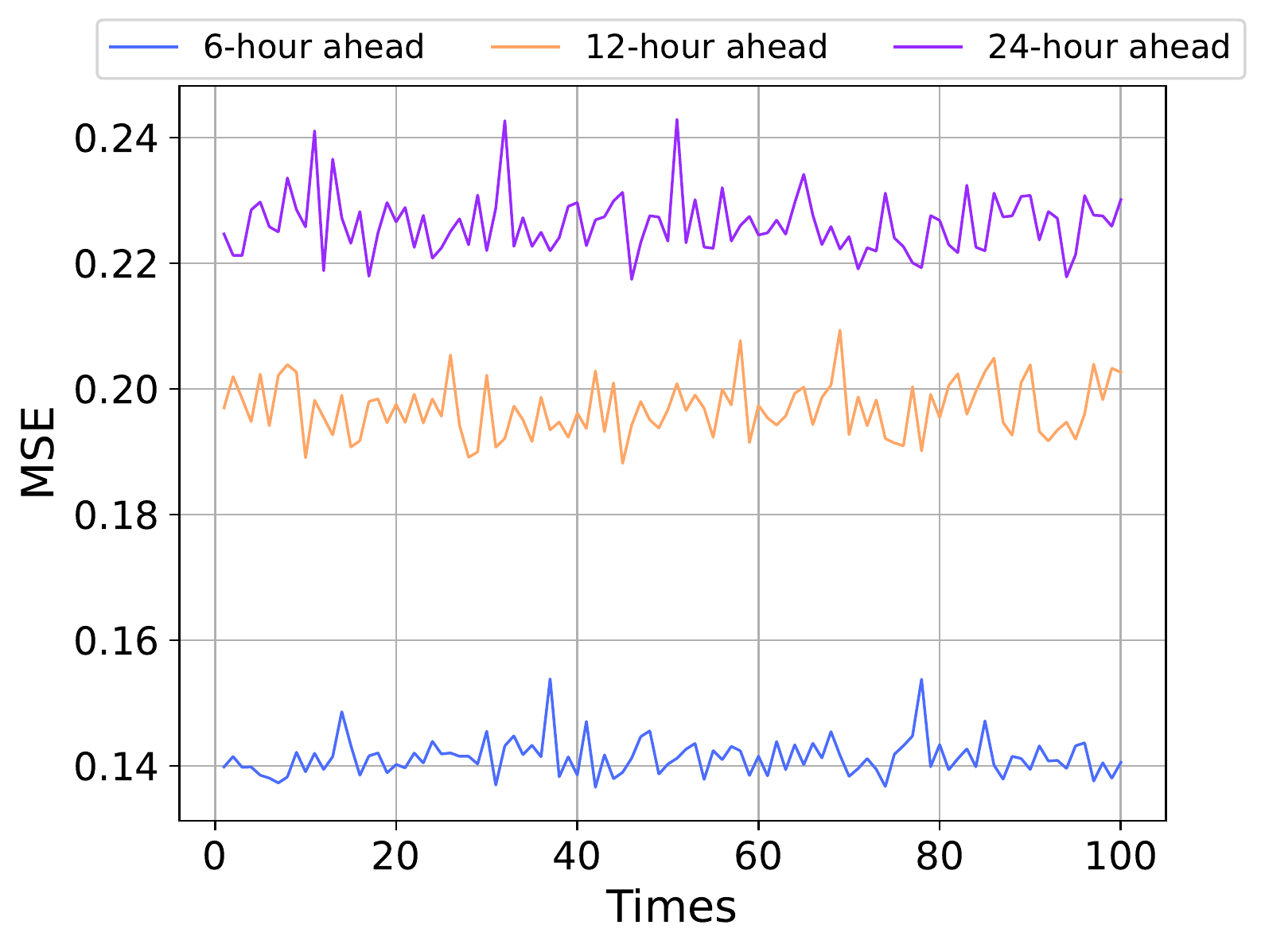}
\label{stability_mse}}
\vspace{-0.0cm}
\subfloat[Distribution of MAE for 100 instances.]{\includegraphics[width=3in]{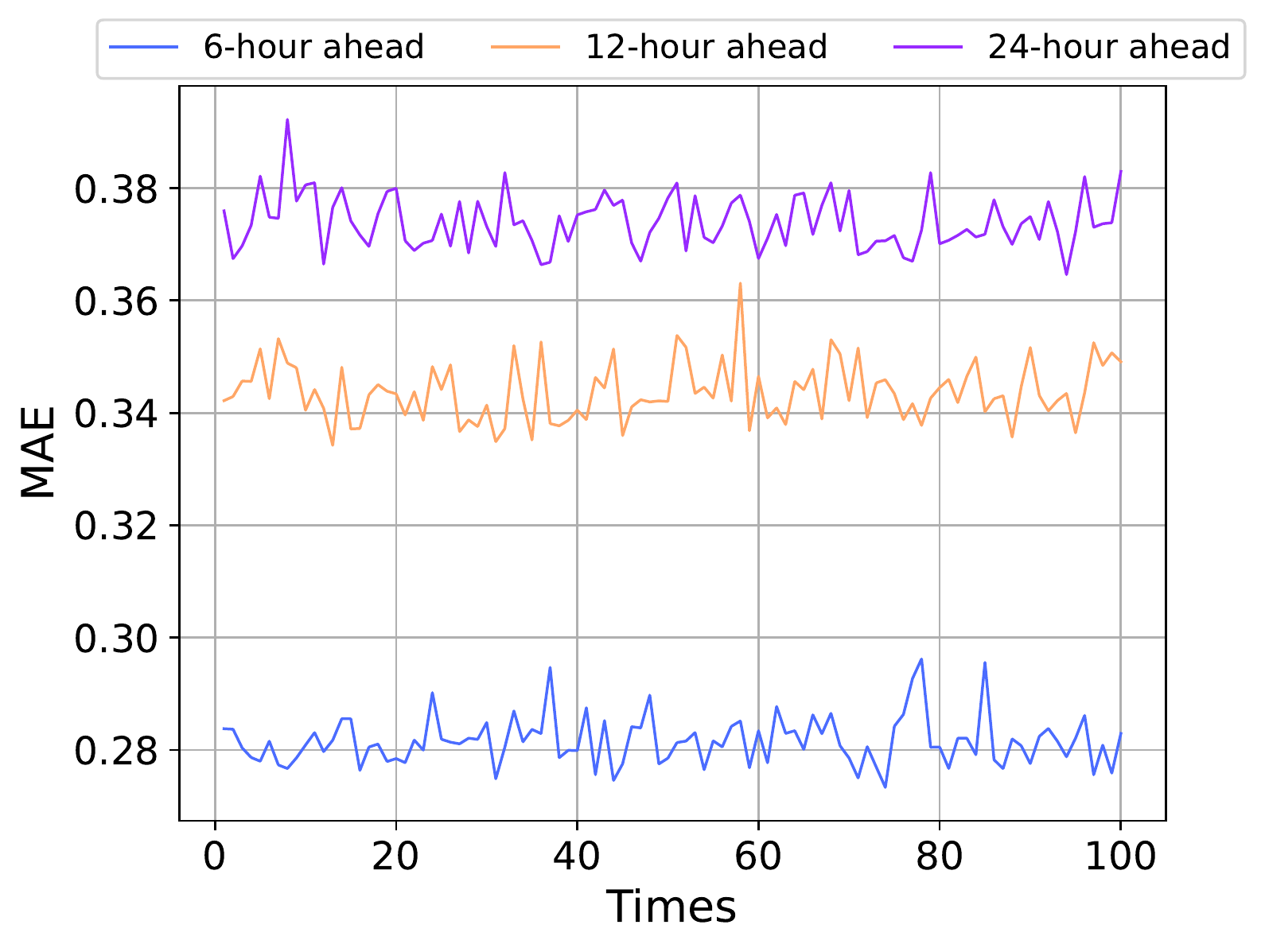}
\label{stability_mae}}
\vspace{-0.0cm}
\caption{Convergence and stability analysis for GFST-WSF on the test
sets.}
\label{stability}
\vspace{-1em}
\end{figure}

\subsection{Convergence and Stability of GFST-WSF}
\indent In order to verify the convergence and stability of GFST-WSF, we re-implemented the training of 100 instances of GFST-WSF. The final distribution of MSE and MAE on the test set are shown in Fig.\ref{stability_mse} and Fig.\ref{stability_mae}, respectively, demonstrating that: the performance of GFST-WSF in terms of convergence and stability is satisfactory, with a small range of fluctuation in both MSE and MAE, as shown in TABLE \ref{stability}. Case studies and convergence analysis reveal that GFST-WSF has good performance and high training stability.

\begin{table}
 \centering
 \caption{Convergence and stability analysis}
 \vspace*{-2mm}
 \label{stability}
 \footnotesize
\begin{tabular}{c|c|c|c|c|c}
\hline \hline
{Lead step} & {Metric} & {Avg} & {Max} & {Min} & {Std.}\\
\cline{1-1}\cline{2-2}\cline{3-3}\cline{4-4}\cline{5-5}\cline{6-6}
\multirow{2}{*}{6-hour} & $\text{MSE}$ & $0.141$ & $0.154$ & $0.136$  & $0.297\%$\\ 
        & $\text{MAE}$& $0.282$ & $0.296$ & $0.273$ & $0.434\%$  \\ 
\cline{1-1}\cline{2-2}\cline{3-3}\cline{4-4}\cline{5-5}\cline{6-6}
\multirow{2}{*}{12-hour} & $\text{MSE}$ & $0.197$ & $0.209$ & $0.188$ & $0.438\%$\\ 
        & $\text{MAE}$& $0.344$ & $0.363$ & $0.334$ & $0.522\%$  \\
\cline{1-1}\cline{2-2}\cline{3-3}\cline{4-4}\cline{5-5}\cline{6-6}
\multirow{2}{*}{24-hour} & $\text{MSE}$ & $0.226$ & $0.243$ & $0.217$  & $0.474\%$\\ 
        & $\text{MAE}$& $0.374$ & $0.392$ & $0.365$ & $0.475\%$  \\
\hline\hline
\end{tabular}
\leftline{}\\
\vspace{-1em}
\end{table}

\subsection{{Summary of Case Studies}}
 \indent In summary of the case studies conducted on the test set, the following phenomena can be observed:\par
\hangafter 1
\hangindent=8.95 mm
\indent 1) LSTM performs poorly compared to deep Transformer-based forecasting models. As the forecasting length increases, the performance of LSTM decreases, while Transformer-based models perform better.\par
\hangafter 1
\hangindent=8.95 mm
\indent 2) FEDformer, which is the latest variant of Transformer, exhibits increasingly significant performance improvements as the forecasting length increases.\par
\hangafter 1
\hangindent=8.95 mm
\indent 3) the introduction of graph neural networks (GNNs) enhances the model's performance compared to methods that only utilize single wind farm data. This is because GNNs can extract spatial features and improve the model's ability to handle uncertainty in wind speed data.\par
\hangafter 1
\hangindent=8.95 mm
\indent 4) GFST-WSF outperforms all benchmark experiments because it not only extracts spatio-temporal features of wind speed data but also performs representation learning in both time and frequency domains.

\section{Conclusion}
\indent In this paper, we proposes a novel GFST-WSF model for short-term wind speed forecasting based on spatio-temporal information. It is the first work to apply Transformer to wind speed forecasting and design a dynamic complex adjacency matrix for GAT. The set of wind farms is modeled as a graph, where nodes with high correlation in wind speed and direction are connected by an edge. The connectivity of nodes in the graph is represented by a complex adjacency matrix, which clearly characterizes the wind speed correlation and time lag between different nodes. GAT can better capture the spatial features of wind speeds based on this matrix. In addition, the variant of Transformer, FEDformer, with its representation learning ability for time series, enables GFST-WSF to better forecast wind speed sequences. The spatio-temporal features obtained by the model can handle noise and uncertainty in wind speed data. The proposed model is verified by comparison with various benchmark testing methods.

\indent Case studies on multi-wind farm data demonstrated the superiority of GFST-WSF in wind speed forecasting. On the testing dataset, compared with state-of-the-art deep learning models, the GFST-WSF achieved a 14\%, 12\%, and 8\% decrease in MSE when forecasting wind speeds for 6, 12, and 24 hours, respectively. These results not only demonstrate the successful application of GFST-WSF in capturing spatial features, but also suggest that deep learning models with deep architectures can better capture complex wind speed variations.


\ifCLASSOPTIONcaptionsoff
  \newpage
\fi



%

%

\bibliographystyle{IEEEtran}%
\bibliography{IEEEabrv,ref5}

\begin{thebibliography}{10}
\providecommand{\url}[1]{#1}
\csname url@samestyle\endcsname
\providecommand{\newblock}{\relax}
\providecommand{\bibinfo}[2]{#2}
\providecommand{\BIBentrySTDinterwordspacing}{\spaceskip=0pt\relax}
\providecommand{\BIBentryALTinterwordstretchfactor}{4}
\providecommand{\BIBentryALTinterwordspacing}{\spaceskip=\fontdimen2\font plus
\BIBentryALTinterwordstretchfactor\fontdimen3\font minus
  \fontdimen4\font\relax}
\providecommand{\BIBforeignlanguage}[2]{{%
\expandafter\ifx\csname l@#1\endcsname\relax
\typeout{** WARNING: IEEEtran.bst: No hyphenation pattern has been}%
\typeout{** loaded for the language `#1'. Using the pattern for}%
\typeout{** the default language instead.}%
\else
\language=\csname l@#1\endcsname
\fi
#2}}
\providecommand{\BIBdecl}{\relax}
\BIBdecl

\bibitem{council2022global}
G.~W.~E. Council, ``Global wind report 2022. 2022,'' \emph{URL https://gwec.
  net/global-wind-report-2022/,(Accessed: 2022-08-17)}.

\bibitem{wind_power_review2011}
X.~Wang, P.~Guo, and X.~Huang, ``A review of wind power forecasting models,''
  \emph{Energy procedia}, vol.~12, pp. 770--778, 2011.

\bibitem{wang2017deep}
H.-z. Wang, G.-q. Li, G.-b. Wang, J.-c. Peng, H.~Jiang, and Y.-t. Liu, ``Deep
  learning based ensemble approach for probabilistic wind power forecasting,''
  \emph{Applied energy}, vol. 188, pp. 56--70, 2017.

\bibitem{wang2021review}
Y.~Wang, R.~Zou, F.~Liu, L.~Zhang, and Q.~Liu, ``A review of wind speed and
  wind power forecasting with deep neural networks,'' \emph{Applied Energy},
  vol. 304, p. 117766, 2021.

\bibitem{2014review}
Y.~Zhang, J.~Wang, and X.~Wang, ``Review on probabilistic forecasting of wind
  power generation,'' \emph{Renewable and Sustainable Energy Reviews}, vol.~32,
  pp. 255--270, 2014.

\bibitem{2020probabilistic}
D.~Nohara, M.~Ohba, T.~Watanabe, and S.~Kadokura, ``Probabilistic wind power
  prediction based on ensemble weather forecasting,'' \emph{IFAC-PapersOnLine},
  vol.~53, no.~2, pp. 12\,151--12\,156, 2020.

\bibitem{NWP}
J.~J. Traiteur, D.~J. Callicutt, M.~Smith, and S.~B. Roy, ``A short-term
  ensemble wind speed forecasting system for wind power applications,''
  \emph{Journal of Applied Meteorology and Climatology}, vol.~51, no.~10, pp.
  1763--1774, 2012.

\bibitem{wang2016analysis}
J.~Wang, Y.~Song, F.~Liu, and R.~Hou, ``Analysis and application of forecasting
  models in wind power integration: A review of multi-step-ahead wind speed
  forecasting models,'' \emph{Renewable and Sustainable Energy Reviews},
  vol.~60, pp. 960--981, 2016.

\bibitem{ARMA}
S.~Rajagopalan and S.~Santoso, ``Wind power forecasting and error analysis
  using the autoregressive moving average modeling,'' in \emph{2009 IEEE power
  \& energy society general meeting}.\hskip 1em plus 0.5em minus 0.4em\relax
  IEEE, 2009, pp. 1--6.

\bibitem{ARIMA}
R.~G. Kavasseri and K.~Seetharaman, ``Day-ahead wind speed forecasting using
  f-arima models,'' \emph{Renewable Energy}, vol.~34, no.~5, pp. 1388--1393,
  2009.

\bibitem{SVR}
K.~Chen and J.~Yu, ``Short-term wind speed prediction using an unscented kalman
  filter based state-space support vector regression approach,'' \emph{Applied
  energy}, vol. 113, pp. 690--705, 2014.

\bibitem{ELM}
X.~Luo, J.~Sun, L.~Wang, W.~Wang, W.~Zhao, J.~Wu, J.-H. Wang, and Z.~Zhang,
  ``Short-term wind speed forecasting via stacked extreme learning machine with
  generalized correntropy,'' \emph{IEEE Transactions on Industrial
  Informatics}, vol.~14, no.~11, pp. 4963--4971, 2018.

\bibitem{2017lightgbm}
G.~Ke, Q.~Meng, T.~Finley, T.~Wang, W.~Chen, W.~Ma, Q.~Ye, and T.-Y. Liu,
  ``Lightgbm: A highly efficient gradient boosting decision tree,''
  \emph{Advances in neural information processing systems}, vol.~30, 2017.

\bibitem{ANN}
E.~Cadenas and W.~Rivera, ``Wind speed forecasting in three different regions
  of mexico, using a hybrid arima--ann model,'' \emph{Renewable Energy},
  vol.~35, no.~12, pp. 2732--2738, 2010.

\bibitem{CNN}
Y.-Y. Hong and T.~R.~A. Satriani, ``Day-ahead spatiotemporal wind speed
  forecasting using robust design-based deep learning neural network,''
  \emph{Energy}, vol. 209, p. 118441, 2020.

\bibitem{RNN}
Q.~Cao, B.~T. Ewing, and M.~A. Thompson, ``Forecasting wind speed with
  recurrent neural networks,'' \emph{European Journal of Operational Research},
  vol. 221, no.~1, pp. 148--154, 2012.

\bibitem{hu2018LSTM}
Y.-L. Hu and L.~Chen, ``A nonlinear hybrid wind speed forecasting model using
  lstm network, hysteretic elm and differential evolution algorithm,''
  \emph{Energy conversion and management}, vol. 173, pp. 123--142, 2018.

\bibitem{hu2019very}
T.~Hu, W.~Wu, Q.~Guo, H.~Sun, L.~Shi, and X.~Shen, ``Very short-term spatial
  and temporal wind power forecasting: A deep learning approach,'' \emph{CSEE
  Journal of Power and Energy Systems}, vol.~6, no.~2, pp. 434--443, 2019.

\bibitem{2014graph-Lstm}
M.~He, L.~Yang, J.~Zhang, and V.~Vittal, ``A spatio-temporal analysis approach
  for short-term forecast of wind farm generation,'' \emph{IEEE Transactions on
  Power Systems}, vol.~29, no.~4, pp. 1611--1622, 2014.

\bibitem{2018CNN-MLP}
Q.~Zhu, J.~Chen, L.~Zhu, X.~Duan, and Y.~Liu, ``Wind speed prediction with
  spatio--temporal correlation: A deep learning approach,'' \emph{Energies},
  vol.~11, no.~4, p. 705, 2018.

\bibitem{2019CNN-LSTM}
Q.~Zhu, J.~Chen, D.~Shi, L.~Zhu, X.~Bai, X.~Duan, and Y.~Liu, ``Learning
  temporal and spatial correlations jointly: A unified framework for wind speed
  prediction,'' \emph{IEEE Transactions on Sustainable Energy}, vol.~11, no.~1,
  pp. 509--523, 2019.

\bibitem{2020GCN-LSTM}
R.~Chen, J.~Liu, F.~Wang, H.~Ren, and Z.~Zhen, ``Graph neural network-based
  wind farm cluster speed prediction,'' in \emph{2020 IEEE 3rd Student
  Conference on Electrical Machines and Systems (SCEMS)}.\hskip 1em plus 0.5em
  minus 0.4em\relax IEEE, 2020, pp. 982--987.

\bibitem{2017GAT}
P.~Velickovic, G.~Cucurull, A.~Casanova, A.~Romero, P.~Lio, Y.~Bengio
  \emph{et~al.}, ``Graph attention networks,'' \emph{stat}, vol. 1050, no.~20,
  pp. 10--48\,550, 2017.

\bibitem{2022transformersSurvey}
Q.~Wen, T.~Zhou, C.~Zhang, W.~Chen, Z.~Ma, J.~Yan, and L.~Sun, ``Transformers
  in time series: A survey,'' \emph{arXiv preprint arXiv:2202.07125}, 2022.

\bibitem{2022FEDformer}
T.~Zhou, Z.~Ma, Q.~Wen, X.~Wang, L.~Sun, and R.~Jin, ``Fedformer: Frequency
  enhanced decomposed transformer for long-term series forecasting,'' in
  \emph{International Conference on Machine Learning}.\hskip 1em plus 0.5em
  minus 0.4em\relax PMLR, 2022, pp. 27\,268--27\,286.

\bibitem{1957cosine_Haversine}
C.~C. Robusto, ``The cosine-haversine formula,'' \emph{The American
  Mathematical Monthly}, vol.~64, no.~1, pp. 38--40, 1957.

\bibitem{2020GNN_review}
J.~Zhou, G.~Cui, S.~Hu, Z.~Zhang, C.~Yang, Z.~Liu, L.~Wang, C.~Li, and M.~Sun,
  ``Graph neural networks: A review of methods and applications,'' \emph{AI
  open}, vol.~1, pp. 57--81, 2020.

\bibitem{2015windtoolkit}
C.~Draxl, A.~Clifton, B.-M. Hodge, and J.~McCaa, ``The wind integration
  national dataset (wind) toolkit,'' \emph{Applied Energy}, vol. 151, pp.
  355--366, 2015.

\end{thebibliography}



\end{document}